\def\year{2019}\relax
\documentclass[letterpaper]{article} 
\usepackage{aaai19}  
\usepackage{times}  
\usepackage{helvet}  
\usepackage{courier}  
\usepackage{url}  
\usepackage{graphicx}  
\usepackage{tabularx}
\usepackage{graphicx}
\usepackage{amsmath}
\usepackage{amssymb}
\usepackage{dsfont}
\usepackage{multirow}
\usepackage[utf8]{inputenc}
\usepackage{array}
\usepackage{algpseudocode}
\usepackage{algorithm}
\usepackage{amsfonts}
\usepackage{enumitem}
\usepackage{multicol}
\usepackage{color}

\newcommand{\fromH}[1]{\textcolor{magenta}{From Huan: #1}}
\newcommand{\nop}[1]{}

\newcommand{\citet}[1]{\citeauthor{#1}~\shortcite{#1}}
\newcommand{\citep}{\cite}

\begin{document}
%
\title{Interactive Semantic Parsing for If-Then Recipes via \\ Hierarchical Reinforcement Learning}

\author{Ziyu Yao$^{\star}$\quad Xiujun Li$^{\dagger}$$^{\S}$\quad Jianfeng Gao$^{\dagger}$\quad Brian Sadler$^{\ddag}$\quad Huan Sun$^{\star}$\\
$^{\star}$The Ohio State University\quad\quad $^{\S}$University of Washington\\
$^{\dagger}$Microsoft Research AI\quad\quad\quad $^{\ddag}$U.S. Army Research Lab\\
{\tt \{yao.470,sun.397\}@osu.edu}\quad\quad {\tt \{xiul,jfgao\}@microsoft.com}\\
{\tt brian.m.sadler6.civ@mail.mil}
}
  
\nop{\author{AAAI Press\\
Association for the Advancement of Artificial Intelligence\\
2275 East Bayshore Road, Suite 160\\
Palo Alto, California 94303\\
}}
\maketitle

\begin{abstract}
Given a text description, most existing semantic parsers synthesize a program in one shot. However, it is quite challenging to produce a correct program solely based on the description, which in reality is often ambiguous or incomplete. In this paper, we investigate \textit{interactive semantic parsing}, where the agent can ask the user clarification questions to resolve ambiguities via a multi-turn dialogue, on an important type of programs called ``If-Then recipes.'' We develop a hierarchical reinforcement learning (HRL) based agent that significantly improves the parsing performance with minimal questions to the user. Results under both simulation and human evaluation show that our agent substantially outperforms non-interactive semantic parsers and rule-based agents.\footnote{All source code and documentations are available at\newline https://github.com/LittleYUYU/Interactive-Semantic-Parsing.} 
\end{abstract}

\section{Introduction}

\begin{table}[ht]
\centering
\small
\begin{tabularx}{\columnwidth}{|X|l}
\hline
\textbf{User}: ``record to evernote'' \\
\textbf{Ground-truth recipe}: [tc: \texttt{Twitter}, tf: \texttt{New liked tweet by you}, ac: \texttt{Evernote}, af: \texttt{Create a link note}]\\
\textbf{\cite{liu2016latent}}: [tc: {\texttt{Phone Call}}, tf: {\texttt{Leave IFTTT any voicemail}}, ac: \texttt{Evernote}, af: {\texttt{Append to note}}]\\
\hline
\hline
\textbf{User}: ``record to evernote'' \\
\textbf{HRL-fixedOrder agent}: ``Which channel triggers the action?'' \\
\textbf{User}: ``twitter''\\
\textbf{HRL-fixedOrder agent}: ``Which event triggers the action?'' \\
\textbf{User}: ``If I like a tweet''\\
\textbf{HRL-fixedOrder agent}: ``Which event results from the trigger?''\\
\textbf{User}: ``Create a note with link''\\
\textbf{Agent Prediction}: [tc: \texttt{Twitter}, tf: \texttt{New liked tweet by you}, ac: \texttt{Evernote}, af: \texttt{Create a link note}] \\
\hline
\hline
\textbf{User}: ``record to evernote'' \\
\textbf{HRL agent}: ``Which event triggers the action?'' \\
\textbf{User}: ``If I like a tweet''\\
\textbf{HRL agent}: ``Which event results from the trigger?''\\
\textbf{User}: ``Create a note with link''\\
\textbf{Agent Prediction}: [tc: \texttt{Twitter}, tf: \texttt{New liked tweet by you}, ac: \texttt{Evernote}, af: \texttt{Create a link note}] \\
\hline
\end{tabularx}
\caption{Semantic parsers on an ambiguous description: The state-of-the-art non-interactive model~\cite{liu2016latent} cannot correctly parse the recipe while our two HRL-based interactive agents can\nop{ parse it correctly}. Particularly, by coordinating the subtask order, the HRL agent asks fewer questions than the HRL-fixedOrder agent (tc: trigger channel, tf: trigger function, ac: action channel: af: action function).} 
\label{tab:motivation_example}
\end{table}

Semantic parsing aims to map natural language to formal domain-specific meaning representations, such as knowledge base or database queries~\cite{berant2013semantic,dong2016language,zhongSeq2SQL2017,gaosurvey}, API calls~\cite{campagna2017almond,su2017building} and general-purpose code snippets~\cite{yin2017syntactic,rabinovich2017abstract}. In this work, we focus on semantic parsing for synthesizing a simple yet important type of conditional statements called \textit{If-Then recipes} (or \textit{If-Then programs}), based on a natural language description~\cite{quirk2015language,beltagy2016improved,liu2016latent,yin2017syntactic,chaurasia2017dialog}. For example, the description ``Create a link note on Evernote for my liked tweets'' should be parsed into an If-Then recipe with 4 components: \textit{trigger channel} \texttt{Twitter}, \textit{trigger function} \texttt{New liked tweet by you}, \textit{action channel} \texttt{Evernote}, and \textit{action function} \texttt{Create a link note}. On the one hand, If-Then recipes allow users to perform a large variety of tasks such as home security (``text me if the door is not locked''). On the other hand, developing intelligent agents that can automatically parse these recipes\nop{or interact with humans when meeting uncertainties represents} is an important step towards complex natural language programming \cite{quirk2015language}. 

Most previous work translates a natural language description to an If-Then recipe \textit{in one turn}: The user gives a recipe description and the system predicts the 4 components. However, in reality, a natural language description can be very noisy and ambiguous, and may not contain enough information. For simplicity, we refer to this problem as \emph{description ambiguity}\nop{the ambiguity of a description}. In fact, in the widely used If-Then evaluation dataset~\cite{quirk2015language}, \textbf{80\%} of the descriptions are ambiguous\nop{\textbf{80\%} of the total $\sim$4K descriptions are considered ambiguous by human annotators}.
As shown in Table~\ref{tab:motivation_example}, the description ``record to evernote'' is paired with the same ground-truth recipe as in the first\nop{aforementioned} example, but even humans cannot tell what the ``record'' refers to (i.e., trigger channel/function) and what kind of note to create on Evernote (i.e., action function). Therefore, it is quite challenging, if not impossible, to produce a correct program {in one shot}\nop{to echo the first sentence} merely based on an ambiguous description. 

Driven by this observation, we investigate \textit{interactive semantic parsing}, where an intelligent agent (e.g., the two HRL-based agents in Table~\ref{tab:motivation_example}) strives to improve the parsing accuracy by asking clarification questions\nop{user a few questions to clarify}. \nop{We address two key challenges in this setting}{Two key challenges are\nop{to be} addressed}: (1) Lack of supervision on when to ask a question. To date, there is no large-scale annotated dataset on whether and when an agent should ask a question during parsing. The only feedback an agent can obtain is whether or not a synthesized program is correct. (2) How to improve the parsing accuracy with a minimal number of \nop{but avoid asking too many} questions? To guarantee a good user experience, the agent should only ask ``necessary'' questions and learn from human interactions over time.

Previous work~\cite{chaurasia2017dialog} developed rule-based agents to interactively predict the 4 components of an If-Then recipe. These agents decide to ask a question when the prediction probability of a recipe component is lower than a predefined\nop{heuristic} threshold. However, such rule-based agents are not trained {in an optimization framework to simultaneously improve the parsing accuracy and reduce the number of questions}.

We address these challenges via a Hierarchical Reinforcement Learning (HRL) approach. We formulate the interactive semantic parsing in the framework of \textit{options} over Markov Decision Processes (MDPs)~\cite{sutton1999between}, where the task of synthesizing an If-Then recipe is naturally decomposed into 4 \textit{subtasks} or \textit{options} (i.e., predicting trigger/action channel/function). In particular, we propose an HRL agent with a hierarchical policy: A high-level policy decides the order of the subtasks to work on, and a low-level policy for each subtask\nop{\footnote{The low-level policy for different subtasks shares the same model structure but different parameters. This is consistent with previous non-interactive work like \cite{liu2016latent} which builds a classifier respectively for each subtask.}} guides its completion by deciding whether to (continue to) ask a clarification question or to predict the subtask component. We train the policies to maximize the parsing accuracy and minimize the number of questions with the rewarding mechanism, where the only supervision (reward signal) is whether or not a predicted component is correct.

\nop{Compared with the widely adopted \textit{flat RL} agent that tries to solve}Compared with the approach of solving the entire task with one flat policy~\cite{mnih2015human}, HRL takes advantage of the naturally defined ``4-subtask'' structure\nop{the structure information (``4-subtask'') which is naturally defined in an If-Then recipe}. Such design also allows the agent to focus on different parts of a recipe description for each subtask, as emphasized \nop{to be critical }in \cite{liu2016latent}, and endows each low-level policy with a reduced state-action space to simplify the learning\nop{each subtask}. On the other hand, the high-level policy optimizes the subtask order by taking into account both the recipe description and user responses. As shown in Table~\ref{tab:motivation_example}, the HRL agent learns to ask about the trigger function first, to which the user response (i.e., ``tweet'') implies the trigger channel. This mechanism leads to fewer questions than the HRL-fixedOrder agent, which executes subtasks in the fixed order of ``tc-tf-ac-af''\nop{(see Table \ref{tab:motivation_example} for notations)}.

Experimental results under both simulation and human evaluation\nop{with both simulated and real user responses} show that our HRL agent can obtain a significantly better accuracy while asking fewer questions than rule-based agents like\nop{as proposed in} \cite{chaurasia2017dialog}. In addition, we show the effectiveness of the high-level policy on reducing the number of questions\nop{that the hierarchical policy can reduce the number of questions by learning the subtask order, and}. Our agent tends to predict\nop{(or work on?)} functions before channels, which is different from most existing works that either assume independence among subtasks {\cite{chaurasia2017dialog}} or predict channels before functions~\cite{beltagy2016improved,dong2016language,yin2017syntactic}.

\section{Background}
If-Then recipes allow people to manage a variety of web services or physical devices and automate event-driven tasks. We focus on recipes from IFTTT.com, where a recipe has 4 components: \textit{trigger channel}, \textit{trigger function}, \textit{action channel}, and \textit{action function}. There are a variety of channels, like \texttt{GMail} and \texttt{Facebook}, representing entities such as web applications and IFTTT-provided services. Each channel has a set of functions representing events (i.e., trigger functions) or action executions (i.e., action functions). {In one recipe, there could be exactly one value for trigger/action channel/function.}

Following~\cite{liu2016latent,chaurasia2017dialog}, we decompose the task of parsing a natural language description into four subtasks, i.e., predicting trigger/action channel/function in a recipe. We have made two key observations about real-life recipe descriptions and existing semantic parsing work: (1) Around 80\% recipe descriptions are ambiguous or contain incomplete information, according to the human annotations provided by~\citet{quirk2015language}, which makes it extremely difficult to synthesize a recipe in one turn (see the prediction result of \cite{liu2016latent} in Table~\ref{tab:motivation_example}). Therefore, \cite{liu2016latent,chaurasia2017dialog,yin2017syntactic} focus on the unambiguous 20\% recipes for evaluation.
(2) Previous work\nop{~\cite{beltagy2016improved,dong2016language,yin2017syntactic,chaurasia2017dialog}} assumes either independence \cite{chaurasia2017dialog} or heuristic dependencies among the 4 subtasks\nop{either independence among the 4 subtasks or some heuristic dependencies among them}. In particular, \citet{liu2016latent} assumes that functions should be predicted before channels since a channel can be derived from the function prediction, while \cite{beltagy2016improved,dong2016language,yin2017syntactic} assume that channels should be predicted before functions and triggers before actions. 

Given these observations, we propose an \textit{interactive semantic parser} that can ask users for clarification to make more accurate predictions. 
Moreover, we {abandon}\nop{do not need} the inter-subtask (in)dependence assumptions used in previous work. Our agent learns to optimize the subtask order\nop{use a learned policy to complete subtasks in an optimized\nop{optimal} order} for each recipe to save questions.

\section{Interactive Semantic Parser}
\subsection{Framework Overview}
\nop{this paragraph tries to explain how we were inspired by previous work, and make a connection to them.} 
Given a recipe description, four components need to be predicted. Thus, the semantic parsing task can be naturally decomposed into four subtasks \nop{$\mathcal{G} = \{g^{tc}, g^{tf}, g^{ac}, g^{af}\}$}{$\mathcal{G} = \{st_{1}, st_{2}, st_{3}, st_{4}\}$}, standing for predicting the trigger channel ({$st_{1}$}), trigger function ({$st_{2}$}), action channel ({$st_{3}$}) and action function ({$st_{4}$}), respectively.
We aim at an agent that can decide the order of subtasks for each parsing task, and only moves to the next subtask when the current one has been completed. We formulate this as a hierarchical decision making problem based on the framework of \textit{options} over Markov Decision Processes (MDPs)~\cite{sutton1999between}. 

Specifically, the agent uses a hierarchical policy consisting of two levels of policies operating at different time scales. The high-level policy selects the next subtask (or \textit{option}) to work on, which can be viewed as operating on a Semi-MDP \cite{sutton1999between}.
The low-level policy selects primitive actions (i.e., predicting a component value or querying the user) to complete the selected subtask. As elaborated in Section~\ref{policy_function}, we adopt 4 low-level policies, each for one subtask. 

\begin{figure}[t!]
\centering
\includegraphics[width=\columnwidth]{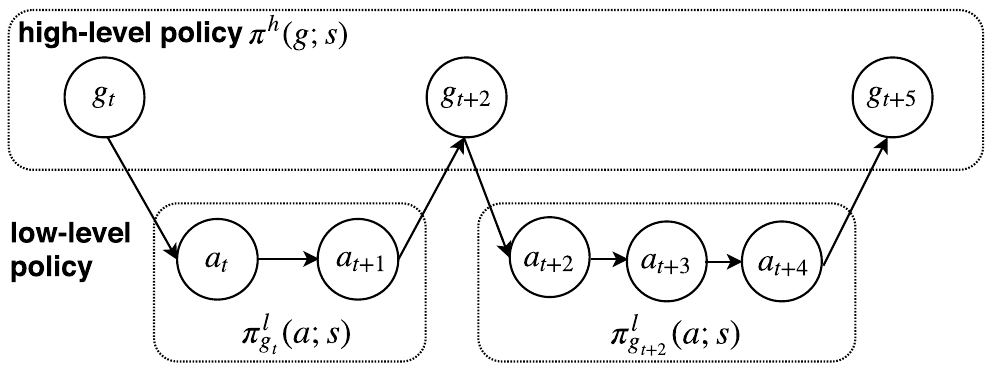}
\caption{Hierarchical policy: The high-level policy chooses a subtask to work on while the low-level policy decides to predict the subtask or ask the user at each time step.}
\label{fig:hrl}
\end{figure}

Figure \ref{fig:hrl} shows the process. At an \textit{eligible} time step $t$ (i.e., \textit{at the beginning of a parsing task or when a subtask terminates}), the high-level policy $\pi^h(g;s_t)$ receives a state $s_t$ and selects a subtask $g_t \in \mathcal{G}$ to work on. Then the low-level policy $\pi_{g_t}^l(a;s_t)$ for this subtask chooses {an} action $a_t \in \mathcal{A}_{g_t}$. By taking this action, the agent receives a low-level reward $r_{g_t}^l(s_t, a_t)$. For the next $N$ time steps ({e.g., $N=2$ in Figure~\ref{fig:hrl}}), the agent will work on the same subtask $g_t$ until it terminates (i.e., when either the agent has predicted the corresponding component or the agent has interacted with the user for $Max\_Lobal\_Turn$ turns).
The agent will receive a high-level reward $r^h(s_t, g_t)$ for this subtask completion, then select the next subtask $g_{t+N}$ and repeat the above procedure until the entire task terminates\nop{is completed} (i.e., when either all four components are predicted or the agent has worked on $Max\_Global\_Turn$ subtasks). 

\noindent \textbf{States.}
A state $s$ tracks 9 items during the course of interactive parsing:
\begin{itemize}[noitemsep,leftmargin=*,topsep=0pt]
\item The initial recipe description $\mathcal{I}$.
\item The boolean indicator $b_i$ $(i=1,2,3,4)$ showing whether subtask $st_i$ has been predicted.
\item The received user answer $d_i$ $(i=1,2,3,4)$ for subtask $st_i$, respectively. \nop{When there is no user answer received, $d_i$ is left empty. Multiple received user answers for the same subtask will be concatenated in order.}
\end{itemize}
For each subtask $st_i$, we learn a \textit{low-level state vector} $s_{st_i}^l$ to summarize state information of this subtask for low-level policy $\pi_{st_i}^l$ to select the next action\nop{ to complete the subtask}. Similarly, a \textit{high-level state vector} $s^h$ is learned to present a summary of the entire state, consisting of the 4 low-level state vectors and other state information, for high-level policy $\pi^h$ to choose the next subtasks. Section~\ref{policy_function} details how states are represented.

\noindent \textbf{Actions.} The action space for the high-level policy is {$\mathcal{G} = \{st_{1}, st_{2}, st_{3}, st_{4}\}$}, where each action denotes one subtask mentioned earlier. 
The action space of subtask $g$ is $\mathcal{A}_g = \mathcal{V}_g \cup \{\text{AskUser}\}$, where $\mathcal{V}_g$ is the set of available component values for subtask $g$ (e.g., all trigger channels for subtask $st_1$), meaning that the agent can either predict a component or ask the user a clarification question.

\noindent \textbf{Rewards.} At each eligible time step $t$, the agent selects a subtask $g_t \sim \pi^h(g;s_t)$ and receives a high-level reward $r^h(s_t,g_t)$ when the subtask terminates after $N$ steps.
The high-level reward will be used to optimize the high-level policy via RL. 
We define $r^h(s_t,g_t)$ as the accumulated low-level rewards from time step $t$ to ${t+N}$. For intermediate time steps (during which the agent works on a selected subtask), there is no high-level reward.
\begin{equation*}
r^h(s_t,g_t) = 
\begin{cases}
	\sum_{k=t}^{t+N}r_{g_t}^l(s_k, a_k) & \text{for eligible } t\\
    0 & \text{otherwise}
\end{cases}
\end{equation*}

While working on subtask $g_t$, the agent receives a low-level reward for taking action $a_t$ (i.e., predicting $g_t$ or querying the user): 
\begin{equation*}
r_{g_t}^l(s_t, a_t) = 
\begin{cases} 
      1 & \text{if } a_t = \ell_{g_t}\\
      -\beta & \text{if } a_t = \text{AskUser}\\
      -1 & \text{otherwise}
\end{cases}
\end{equation*} 
where $\ell_{g_t}$ is the ground-truth label for subtask $g_t$ and $\beta \in [0,1)$ is the penalty for querying the user. The received reward will be used to optimize the low-level policy $\pi_{g_t}^l$ for this subtask via RL.

Essentially, the low-level reward $r_{g}^l$ alleviates the reward sparsity in the long trajectory of the entire task, and stimulates the agent to predict a correct component with fewer questions.
Note that during the course of RL we do not stop the parsing even if one of the predictions is incorrect in order to encourage the agent to predict as more correct components as possible. This also fits the realistic application setting where the agent does not know the ground truth at the component level, and does not receive the external reward signal until it recommends a synthesized program to the user at the end of the interactive parsing process.

\noindent \textbf{Transition.} Interactive semantic parsing starts with a state $s_0$ where $b_i=0$ (i.e., no subtask is completed) and $d_i=\varnothing$ (i.e., no user answer). As the agent takes actions, it deterministically transits to a new state with updated $b_i$ and $d_i$.

\subsection{Hierarchical Policy Functions}
\label{policy_function}
\paragraph{Low-level policy function.} The low-level policy {$\pi_{st_i}^l(a;s)$} decides whether to ask a question or predict subtask {$st_i$} (e.g., {selecting a trigger channel name for $st_1$}). When it selects the ``AskUser'' action, the user will clarify the subtask with a natural language utterance as shown in Table \ref{tab:motivation_example}. The policy then decides the next action by considering both the recipe description and the user response.

\begin{figure}[t!]
\centering
\includegraphics[height=6.5cm]{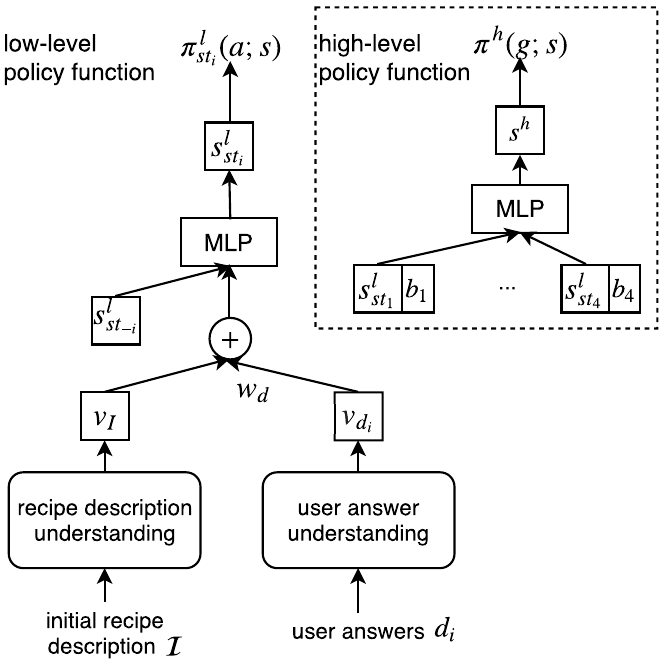}
\caption{High- and low-level policy designs.}
\label{fig:policy_function}
\end{figure}

To understand a recipe description, we choose one of the state-of-the-art models, Latent Attention Model (LAM) \cite{liu2016latent}. The main idea behind LAM is to first understand the latent sentence structure and then pay attention to words that are critical for a subtask. For example, given a description with a pattern ``X to Y,'' LAM adopts a latent attention mechanism to first locate the keyword ``to,'' and then focus more on ``X'' when predicting the trigger channel/function and on ``Y'' when predicting the action channel/function. This reveals that different subtasks need different policies, so that they can focus on different parts of a recipe description. Such design also allows each subtask to have a different and reduced action space. Hence, we define 4 low-level policy functions with the same model structure yet different parameters for the 4 subtasks, respectively.

To deal with user responses, one straightforward adaptation from LAM is to simply concatenate them with the initial recipe description as the input and extend the prediction space with an extra ``AskUser'' action. However, we observe that user answers are fundamentally different from a recipe description, e.g., user answers typically contain information related to the queried subtask (rather than all subtasks). Therefore, we propose to model these two parts separately as shown in Figure \ref{fig:policy_function}. In particular, we employ LAM to instantiate the ``\textit{recipe description understanding}'' module to capture the meaning of the initial recipe description (i.e., $v_I$), and utilize a bidirectional GRU-RNN with the attention mechanism~\cite{zhou2016attention} to instantiate the ``\textit{user answer understanding}'' module.\footnote{We also verified this separate design is much better than the straightforward adaptation from LAM mentioned earlier during model development.} Details can be found in the Appendix \ref{app:user_ans_understand}.
\nop{Specifically, a trainable vector $c$ is learned to calculate the attention weight for the $j$-th word in the user answer $d_i$, i.e., $w_{att_j} = \text{softmax}(c^Th_{d_{i, j}})$, where $h_{d_{i,j}} = [\overrightarrow{h}_{d_{i, j}}; \overleftarrow{h}_{d_{i, j}}]$ is a concatenation of the forward and backward RNN hidden states for the word, and $d_i$ is finally represented as $v_{d_i} = \sum_{j}w_{\text{att}_j}h_{d_{i,j}}$.}
{Since each subtask maintains a separate policy function, the learned $v_I$ can be different for different subtasks. The agent can ask the user to clarify a subtask for multiple times,\footnote{Using the same predefined question (see User Simulation\nop{Section~\ref{usersimulation}}).} and the newly received answers will be concatenated with the old ones to update $d_i$.}

We combine the representations of recipe description ($v_I$) and user answers ($v_{d_i}$) as the semantic representation related to subtask $st_i$:
\begin{equation}
\label{eq:1}
{v_i} = (1-w_d) v_I + w_d v_{d_i},
\end{equation}
\nop{should we associate the representation with index i?}where $w_d \in [0, 1]$ is a weight controlling information from the initial recipe description versus from user answers. {The semantic vector $v_i$, concatenated with the information of other subtasks, defines the \textit{low-level state vector} $s_{st_i}^l$ of subtask $st_i$ via a multi-layer perceptron model (the ``{MLP}'' module in Figure \ref{fig:policy_function}):}
\begin{equation}
\label{eq:2}
\small
s_{st_i}^l = \text{tanh}(W_{c_i} [s_{st_1}^l; ...; s_{st_{i-1}}^l; v_i; s_{st_{i+1}}^l; ...; s_{st_4}^l]).\footnote{Bias terms are omitted for clarity.}
\end{equation}

Essentially, $s_{st_i}^l$ summarizes the state information for completing subtask $st_i$, including the initial recipe description, user answers for $st_i$, and the current low-level state vectors of other subtasks (such that the completion status of other subtasks can affect the current one, as shown in Table \ref{tab:motivation_example}). Finally, the low-level policy function $\pi_{st_i}^l(a;s)$$=$$\text{softmax}(W_{st_i}^l s_{st_i}^l)$, takes the state vector as the input, and outputs a probability distribution over the action space of subtask $st_i$.
\nop{All low-level policies are stochastic in that the next action is sampled according to the probability distribution to allow exploration in RL.}

\noindent \textbf{High-level policy function.}
The high-level policy $\pi^h(g;s)$ receives a state $s$ and decides the next subtask $g$. The \textit{high-level state vector} $s^h$ is learned to encode the state of overall parsing task through a multi-layer perceptron model (i.e., the ``{MLP}'' module in Figure~\ref{fig:policy_function}) using the 4 low-level state vectors $s_{st_i}^l$'s, as well as the subtasks' boolean indicators $b_i$'s, as inputs:
\begin{gather}
s^h = \text{tanh}(W_c [s_{st_1}^l; b_1; s_{st_2}^l; b_2; s_{st_3}^l; b_3; s_{st_4}^l; b_4]), \nonumber \\
\pi^h(g;s) = \text{softmax}(W^h s^h). \label{eq:3}
\end{gather}

\noindent \textbf{Optimization.}
The high-level policy $\pi^h$ is trained to maximize the expectation of the discounted cumulative rewards for selecting subtask $g_t$ in state $s_t$:
{\small
\begin{align*}
&\max_{\pi^h} J(\theta) = \max_{\pi^h} E_{\pi^h} [r^h(s_t, g_t) \stepcounter{equation}\tag{\theequation}\label{eq:4}\\
&+ \gamma r^h(s_{t+N_1}, g_{t+N_1}) + \gamma^2 r^h(s_{t+N_1+N_2}, g_{t+N_1+N_2}) \\
&+ ... + \gamma^\infty r^h(s_{t+\sum_{n=1}^\infty N_n}, g_{t+\sum_{n=1}^\infty N_n}) | s_t, g_t, \pi^h(\theta)],
\end{align*}
}
\noindent where $\theta$ stands for parameters in $\pi^h$, $N_n (n=1,2,...,\infty)$ is the number of time steps that the agent spent on the previous subtask\nop{what do you mean by the last subtask?}, and $\gamma \in [0, 1]$ is the discount factor\nop{controlling how much an action depends on long-term rewards}.
Similarly, we train\nop{Similar to learning the high-level policy, we train} the low-level policy $\pi_{g_{t}}^l$ for the selected subtask $g_t$ to maximize its expected cumulative discounted low-level reward: 
{
\begin{align*}
&\max_{\pi_{g_t}^l}J_{g_t}(\phi_{g_t}) = \max_{\pi_{g_t}^l} E_{\pi_{g_t}^l}[ \stepcounter{equation}\tag{\theequation}\label{eq:5}\\
&\sum_{k \geq 0} \gamma^k r_{g_t}^l(s_{t+k},a_{t+k}) | s_t, a_t, g_{t}, \pi_{g_t}^l(\phi_{g_t})],
\end{align*}
}
\noindent where $\phi_{g_t}$ denotes the parameters in $\pi_{g_t}^l$, and $\gamma$ is the same discount factor.\nop{as that of high level.}

All policies are stochastic in that the next subtask or action is sampled according to the probability distribution which allows exploration in RL, and that the policies can be optimized using policy gradient methods.\nop{All policies are stochastic, and thus can be optimized using policy gradient methods.}
In our experiments we used the REINFORCE algorithm~\cite{williams1992simple}. Details are outlined in Algorithm~\ref{alg:training} of the {Appendix.}

\section{Experiments}
\label{sec:experiments}
We experiment with our proposed HRL agent under both simulation and human evaluation\nop{using both simulated and real user responses}. 

\begin{table}[t!]
\centering
\begin{tabular}{c|cccc}
\hline
\multirow{2}{*}{Test Data} & \multirow{2}{*}{CI} & \multicolumn{2}{c}{VI} & \multirow{2}{*}{Total} \\
\cline{3-4} & & VI-1/2 & VI-3/4 & \\
\hline
Size & 727 & 1,271 & 1,872 & 3,870 \\
(\%) & (18.79) & (32.84) & (48.37) & (100) \\
\hline
\end{tabular}
\caption{Statistics of the test subsets.}
\label{tab:dataset}
\end{table}

\subsection{Dataset}
We utilize the 291,285 $<$recipe, description$>$ pairs collected by~\citet{ur2016trigger} for training and the 3,870 pairs from \citet{quirk2015language} for testing.\footnote{\citet{quirk2015language} only released the urls of recipes in their test set, among which the unavailable ones have been removed from our test set. Each recipe is {associated with} a unique ID. We ensure no overlapping recipes between training and testing set by examining their IDs. \nop{what do you mean by unavailable recipes? How do you define overlapping (or maybe change it to another word to be succinct?)}} 20\% of the training data are randomly sampled as a validation set. All recipes were created by real users on IFTTT.com. In total, the datasets involve 251 trigger channels, 876 trigger functions, 218 action channels and 458 action functions. For each description in the test set, \citet{quirk2015language} collected five recipe annotations from Amazon Mechanical Turkers. For each subtask, if at least three annotators make the same annotation as the ground truth, we consider this recipe description as \textit{clear} for this subtask\nop{, called ``clear subtask (CI)'';}; otherwise, it is labeled as \textit{vague}\nop{\textit{vague} description}\nop{, called ``vague subtask (VI)''}. In this way, we split the entire test set into three subsets as shown in Table~\ref{tab:dataset}: (1) \textit{CI}: 727 recipes whose descriptions are clear for all 4 subtasks; (2) \textit{VI-1/2}: 1,271 recipes containing 1 or 2 vague subtasks; (3) \textit{VI-3/4}: 1,872 recipes containing 3 or 4 vague subtasks.

\subsection{Methods Comparison}
\begin{itemize}[noitemsep,leftmargin=*,topsep=0pt]
\item \textbf{LAM}: The Latent Attention Model~\cite{liu2016latent},\footnote{{Unlike \cite{liu2016latent}, which consolidates channel and function names (e.g., ``\texttt{Twitter.New liked tweet by you}'') and builds 2 classifiers for trigger and action respectively, we develop 4 classifiers so that an agent can inquire channel or function separately.}} one of the state-of-the-art models for synthesizing If-Then recipes. We do not consider the model ensemble in~\cite{liu2016latent}, as it can be applied to all other methods as well. Our reproduced LAM obtains a performance close to the reported one without ensemble.
\item \textbf{LAM-rule Agent}: A rule-based agent built on LAM, similar to \cite{chaurasia2017dialog}. Specifically, the agent makes a prediction on a subtask with a certain probability. If the probability is lower than a threshold,\footnote{We set it at 0.85 based on validation set.} the user is asked a question. The user answer is concatenated with the initial recipe description for making a new prediction. This procedure repeats until the prediction probability is greater than the threshold or the agent has run for $Max\_Local\_Turn$ turns\nop{has asked $Max\_Local\_Turn$ questions} on the subtask. 
\item \textbf{LAM-sup Agent}: {An agent based on LAM, but with the user answer understanding module in Figure \ref{fig:policy_function} and an extra ``AskUser'' action for each subtask}\nop{We design \textbf{LAM-sup} with the same policy function and action space (i.e., with an ``AskUser'' action) as the low-level policy in our HRL agent (see Figure \ref{fig:policy_function})}. It is trained via a supervised learning strategy and thus is named LAM-sup\nop{train it with a supervised learning strategy (hence the name LAM-sup)}. We collected the training data for each subtask $st_i$ {based on the LAM-rule agent}: If LAM-rule completes the subtask without interactions with humans, we add a tuple $<$$\mathcal{I}, \varnothing, \ell_{st_i}$$>$ to the training set, where $\mathcal{I}$ is the recipe description and $\ell_{st_i}$ is the ground-truth label for subtask $st_i$; otherwise, we add two tuples $<$$\mathcal{I}, \varnothing, \text{``AskUser''}$$>$ and $<$$\mathcal{I}, d_i, \ell_{st_i}$$>$, where $d_i$ is the received user answer. We train the agent by minimizing the cross entropy loss. During testing, for each recipe description, the agent starts with no user answer; for each subtask, if it predicts the ``AskUser'' label, the received user answer will be {concatenated with previous ones} to make a new prediction until a non-AskUser label is selected.
\item \textbf{HRL Agent}: Our agent with a two-level hierarchical policy.\nop{in which the high-level policy controls the executing order of the subtask, and the low-level policy is to accomplished the selected subtask.}
\item \textbf{HRL-fixedOrder Agent}: A variant of our HRL agent with a fixed subtask order of ``$st_{1}$-$st_{2}$-$st_{3}$-$st_{4}$'' and no high-level policy learning.
\end{itemize}

\begin{table*}[!t]
\centering
\begin{tabular}{>{\centering\arraybackslash}p{7em}| >{\centering\arraybackslash}p{2.5em} | >{\centering\arraybackslash}p{2.5em}| >{\centering\arraybackslash}p{2.5em}|>{\centering\arraybackslash}p{2.5em}| >{\centering\arraybackslash}p{2.5em}| >{\centering\arraybackslash}p{2.5em}| >{\centering\arraybackslash}p{2.5em}| >{\centering\arraybackslash}p{2.5em}| >{\centering\arraybackslash}p{2.5em}| >{\centering\arraybackslash}p{2.5em}| >{\centering\arraybackslash}p{2.5em}| >{\centering\arraybackslash}p{2.5em}| >{\centering\arraybackslash}p{2.5em}}
\cline{2-12} & \multicolumn{9}{c|}{{Simulation Eval}} & \multicolumn{2}{c|}{{Human Eval}}\\\hline 
\multicolumn{1}{|c|}{\multirow{2}{4em}{Model}} & \multicolumn{3}{c|}{All} & \multicolumn{2}{c|}{CI} & \multicolumn{2}{c|}{VI-$1/2$} & \multicolumn{2}{c|}{VI-$3/4$} & \multicolumn{2}{c|}{VI-$3/4$} \\\cline{2-12} 
\multicolumn{1}{|c|}{} & C+F Acc & Overall Acc & $\#$Asks & C+F Acc & $\#$Asks & C+F Acc & $\#$Asks & C+F Acc & $\#$Asks & C+F Acc & \#Asks \\
\hline 
\hline
  \multicolumn{1}{|c|}{LAM} & 0.374 & 0.640 & 0 & 0.801 & 0 & 0.436 & 0 & 0.166 & 0 & 0.206 & 0\\
\hline
\hline
 \multicolumn{1}{|c|}{LAM-rule} & 0.761 & 0.926 & 3.891 & 0.897 & 1.433 & 0.743 & 2.826 & 0.721 & 5.568 & 0.518 & 2.781 \\
 \multicolumn{1}{|c|}{LAM-sup} & 0.809 & 0.940 & \textbf{2.028} & 0.894 & \textbf{0.684} & 0.803 & \textbf{1.482} & 0.780 & 2.921 & 0.433 & 2.614 \\
\hline
\multicolumn{1}{|c|}{HRL-fixedOrder} & 0.881 & {0.966} & 2.272 & \textbf{0.950} & 1.522 & 0.855 & 1.958 & 0.871 & 2.777 & 0.581 & 2.306$^*$\\
\multicolumn{1}{|c|}{HRL} & \textbf{0.894}$^*$ & \textbf{0.968} & 2.069$^*$ & 0.949 & 1.226$^*$ & \textbf{0.888}$^*$ & 1.748$^*$ & \textbf{0.878}$^*$ & \textbf{2.615}$^*$ & \textbf{0.634$^*$} & \textbf{2.221$^*$}\\
  \hline
\end{tabular}
\caption[]{Model evaluation on the test set. For Simulation Eval, each number is averaged over 10 runs. For Human Eval, the LAM result is calculated on the sampled 496 recipes. $^*$ denotes significant difference in mean between HRL-fixedOrder vs. HRL in Simulation Eval\nop{($p < 0.01$)} and between HRL-based agents vs. \{LAM-rule, LAM-sup\} agents in Human Eval ($p < 0.05$).}
\label{tab:test}
\end{table*}

\noindent \textbf{Evaluation metrics.} 
We compare each method on three metrics: (1) \textit{C+F Accuracy}: A recipe is considered parsed correctly only when all its 4 components (i.e., \underline{C}hannel+\underline{F}unction) are accurately predicted, as {adopted in \cite{quirk2015language,liu2016latent}}. (2) \textit{Overall Accuracy}: We measure the average correctness of predicting 4 components of a recipe, e.g., the overall accuracy for predicting 3 components correctly and 1 incorrectly is 0.75. (3) \textit{\#Asks}: The averaged number of questions for completing an entire task. Generally, the \textit{C+F Accuracy} is more challenging as it requires no mistake on any subtask. On the other hand, \textit{\#Asks} can reveal if an agent asks redundant questions. In our experiments, we consider \textit{C+F Accuracy} and \textit{\#Asks} as two primary metrics.
\nop{I think comparing channel accuracy is trivial since predicting channels is usually simpler, so we don't compare it.}

\noindent \textbf{Implementation details.} 
The word vector dimension is set at 50, the weight factor $w_d$ is $0.5$, and the discount factor $\gamma$ is 0.99. {The max turn $Max\_Local\_Turn$ is set at 5 and\nop{the max subtask turns} $Max\_Global\_Turn$ at 4}, which allows four subtasks at most. \nop{To accelerate the training of HRL-based agents, we adopt a similar skill as \cite{gordon2017iqa,wang2017video} to initialize their low-level policy functions with the trained LAM-sup model. }$\beta$ is a trade-off between parsing accuracy and the number of questions: A larger $\beta$ trains an agent to ask fewer questions but with less accuracy, while a lower $\beta$ leads to more questions and likely more accurate parses. With the validation set, we experimented with $\beta$$=$$\{0.3,0.4,0.5\}$, and observed that when $\beta=0.3$, the number of questions raised by the HRL-based agents is still reasonable compared with LAM-rule/sup, and its parsing accuracy is much higher. More details are shown in Appendix \ref{app:imple_detail}.

\nop{To accelerate the training of HRL-based agents, we adopt a similar skill as \cite{gordon2017iqa,wang2017video} initialize their low-level policy functions with the well-trained LAM-sup model. Particularly for HRL, we apply an ``action mask'' at high level, so that the completed subtask will not be repeatedly chosen as the next subtask. However, the agent does not rely on this mask after being fully trained. The validation check is performed every 2K iterations on a randomly sampled 1K validation recipes. The best parameter setting is the one that obtains the most high-level reward in validation.}

\subsection{User Simulation}
\label{usersimulation}
In our work, for each subtask, agent questions are predefined based on templates,\footnote{Automatically generating recipe-specific questions is non-trivial, which we leave for future work.} and a user answer is a natural language description about the queried subtask. Given that it is too costly to involve humans in the training process, we introduce a user simulator to provide answers to agent questions.

For a trigger/action function, we adopt several strategies to simulate user answers, including revising its official description on IFTTT.com and replacing words and phrases in the function description with their paraphrases according to the PPDB paraphrase database~\cite{pavlick2015ppdb}. In addition, we extract user descriptions of a function from our training set, based on six manually defined templates. For example, for a recipe description following pattern ``If X then Y,'' X will be considered as an answer to questions about the ground-truth trigger function and Y as an answer to those about the ground-truth action function. 
{More details regarding the strategies are in Appendix\nop{~\ref{app:user_sim}} and source code will also be released.}
For each function, we collected around 20 simulated user answers on average. 
In our simulations, for each question we randomly select one from this set of possible answers as a response.

For trigger/action channels, when an agent asks a question, the user simulator will simply provide the channel name (e.g., \texttt{GMail}), since it is straightforward and natural for real users as well.
 
\subsection{Simulation Evaluation}
Table \ref{tab:test} shows results on the test set in the simulation environment, where our user simulator provides an answer when requested\nop{try to be succinct}\nop{ and $^*$ denotes significant difference in mean between HRL-fixedOrder vs. HRL ($p < 0.01$)}. By enabling the user to clarify, all\nop{interactive} agents obtain much better accuracy compared with the original {non-interactive} LAM model. In particular, HRL-based agents outperform others by \textbf{7\% $\sim$ 13\%} on C+F accuracy and \textbf{2\% $\sim$ 4\%} in terms of Overall accuracy. For vague recipes in VI-1/2 and VI-3/4 subsets, which make up more than 80\% of the entire test set, the advantage of HRL-based agents is more prominent. For example, on VI-3/4, HRL-based agents obtain 9\% $\sim$ 15\% better C+F accuracy than LAM-rule/sup, yet with fewer questions, indicating that they are much more able to handle ambiguous recipe descriptions. 

Compared with HRL-based agents, the LAM-rule agent usually asks the most questions, partly because it relies on heuristic thresholding to make decisions. On VI-3/4, it asks twice the number of questions but parses with 15\% less accuracy than HRL-based agents. On the other hand, the LAM-sup agent always asks the least questions, especially when the recipe description is relatively clear (i.e., CI and VI-1/2). However, it may simply miss many necessary questions, leading to at least 5\% accuracy loss. 

Finally, we evaluate the high-level policy by comparing HRL with HRL-fixedOrder. The significance test shows that \nop{HRL asks much fewer questions, although predicts with a close or slightly better accuracy}HRL requires fewer questions to obtain a similar or better accuracy. Interestingly, we observe that, under the interactive environment, HRL tends to predict the function before the channel, which is different from the inter-task (in)dependence assumptions in previous work. This is mainly because users' descriptions of a function can be more specific and may contain information about its channel. The HRL agent is thus trained to utilize this intuition for asking fewer questions. 

\nop{\begin{table}[t]
\centering
\begin{tabular}{|c|c|c|}
\hline
Model &  C+F Acc & \#Asks\\
\hline
\hline
LAM & 0.206 & 0 \\
\hline
\hline
LAM-rule & 0.518 & 2.781 \\
LAM-sup & 0.433 & 2.614 \\
\hline
HRL-fixedOrder & 0.581 & 2.306$^*$ \\
HRL & \textbf{0.634$^*$} & \textbf{2.221$^*$} \\
\hline
\end{tabular}
\caption{\nop{Model evaluation with {real user responses}}Human evaluation on the VI-3/4 subset.{ The result of LAM is calculated on the sampled 496 recipes.} $^*$ denotes significant difference in mean ($p < 0.05$) between HRL-based agents and \{LAM-rule, LAM-sup\} agents. \nop{($^*$: sig. diff. from HRL-fixedOrder, see Table~\ref{tab:test})}\nop{$^*$ denotes significant difference in mean ($p < 0.01$) between HRL-fixedOrder and HRL.}\nop{To ensure user experience and for easier comparison, each agent is limited to ask at most one question for each subtask \fromH{remove this sentence? as it appears in main content. Also you may add one sentence about not showing overall ACC anymore}.}}
\label{tab:userstudy}
\end{table}}

\subsection{Human Evaluation}
\label{human_eval}
We further conduct human evaluation \nop{on the four interactive agents. In particular, we test them}to test the four interactive agents on the most challenging VI-3/4 subset. Two students familiar with IFTTT were instructed to participate in the test. For each session, a recipe from VI-3/4\nop{ was randomly selected,} and one agent from \{LAM-rule, LAM-sup, HRL-fixedOrder, HRL\} \nop{was}were randomly picked. The participants were presented the description and the ground-truth program components of the recipe, and were instructed to answer clarification questions prompted by the agent with a natural language sentence. To help the participants better understand the recipe, we also showed them the official explanation of each program component. However, we always encouraged them to describe a component in their own words when being asked\nop{express the component in their own words and to avoid using the same phrase from the explanation}. For a better user experience and an easier comparison, we limited each agent to ask at most one question for each subtask. In total, we collected 496 conversations between real humans and the four agents. {Examples are shown in Appendix \ref{app:examples}.}

We compare each agent primarily on \textit{C+F Accuracy} and \textit{\#Asks}. As shown in Table \ref{tab:test}\nop{\ref{tab:userstudy}},
\nop{In general, all agents perform much better than the non-interactive LAM model, though worse than the simulation results, mainly because they are trained with the user simulator, and the language complexity of {real users}\nop{human annotators} is higher (e.g., the Out-Of-Vocabulary issue). How to simulate user responses as close as possible to real ones for training is a non-trivial task, which we leave for future work.}all agents perform much better than the non-interactive LAM model. Particularly, the HRL agent outperforms the LAM model by $>$40\% accuracy, with an average of $\sim$2.2 questions on VI-3/4 (which {is a reasonable number of questions} as each task contains at least 3 vague subtasks). We also observe that the two HRL-based agents obtain 6\% $\sim$ 20\% better parsing accuracy \nop{($>$15\% accuracy) }with even fewer questions than the LAM-rule/sup agents.\nop{ $^*$ denotes significant difference in mean between HRL-based agents vs. LAM-rule/sup agents ($p < 0.05$).} Moreover, in comparison with the HRL-fixedOrder agent, the HRL agent can synthesize programs with a much better accuracy but fewer questions, showing the benefit of optimizing subtask order at the high level. However, there is still large space to improve compared with simulation results in Table~\ref{tab:test}, mainly because agents are trained with the user simulator while the language {used by} real users for answers can be very different ({e.g., having the Out-Of-Vocabulary issue, misspellings, less or non- relevant information}). How to simulate user responses as close as possible to real ones for training is a non-trivial task, which we leave for future work.


\subsection{Discussion}
Here we further discuss our framework and future work on the following aspects:

\noindent\textbf{Error analysis.} Due to the discrepancy between real users and simulated users, several major factors affect the performance of the HRL agent, including user typos (e.g., ``emial'' for ``email'') and unseen expressions (e.g., ``i tweeted something'' to describe the function \texttt{New tweet by you})\nop{, and confusion between similar functions}. To improve the robustness of the HRL agent, possible solutions as future work can be modeling user noises in simulation~\cite{li2016user}, or crowdsourcing more diverse component descriptions {\nop{(which can be very costly) }as user answers for training}\nop{or\nop{and another solution is} to fine-tune the agent with real users via RL, which we leave for future work}.

\noindent \textbf{Training with real users in the loop.} {Theoretically, our agents can be trained with real users. However, it is too costly to be practical because the agent can require many interactions during the training phase}. An alternative way is to train the agent in simulation and fine-tune it with real users, or to build a \textit{world model} that mimics real user behaviors during the human-in-the-loop training \cite{peng2018deep}. Both approaches need significant efforts to be carefully designed, which we leave as future work.

\noindent \textbf{Generalizability to other semantic parsing tasks.} The proposed HRL framework can be easily generalized to resolve ambiguities in other semantic parsing tasks where subtasks can be pre-defined. For example, in the knowledge-graph-based question answering task \cite{berant2013semantic,yih2015semantic}, the subtasks include identifying entities, predicting relations, and associating constraints. To train the HRL agent, one can build a user simulator by paraphrasing the ground-truth entities or relations, {similar to Section \ref{usersimulation}.} HRL can be very promising for these tasks, as it enables temporal abstractions over the state and action space (leading to a smaller search space) and can model the dependencies between subtasks, as shown in Table \ref{tab:motivation_example}. We will explore these applications in the future. 

\section{Related Work}
\nop{In addition to previous work about If-Then program synthesis~\cite{quirk2015language,beltagy2016improved,liu2016latent, dong2016language,yin2017syntactic}} 
In addition to the aforementioned work on If-Then program synthesis, \citet{confidence} investigated how to measure a semantic parser's confidence in its predictions, but did not further resolve uncertainties. Others include\nop{Other related work includes\nop{our work is also related to}}:

\noindent \textbf{Interactive Systems for Resolving Ambiguities.} Resolving ambiguities via interactions with humans has been explored in \nop{areas including }Natural Language Understanding in dialog systems \cite{thomason2015learning,dhingra2017towards}, Question Answering \cite{guo2016learning,li2017context}, CCG parsing \cite{he2016human} and parsing for SQL and web APIs \cite{li2014constructing,gur2018dialsql,su2018natural}.
\citet{guo2016learning} built an agent to ask relevant questions until it has enough information to correctly answer user’s question, but expected the user to respond with an oracle value. \citet{he2016human} investigated generating multi-choice questions for humans to resolve uncertainties in parsing sentences. They determined the necessity of a question by a heuristic threshold. In contrast, we allow users to respond with natural language utterances, and our HRL-based agents can learn when to ask through {the reward mechanism}. 
Recently, \cite{li2016dialogue,azaria2016instructable,iyer2017learning} explored human feedback {on the final results} as training supervision. Different from theirs, we include humans during the parsing process for them to provide necessary information in natural language, and define rewards as the only feedback.

\noindent \textbf{Hierarchical Reinforcement Learning (HRL).} To solve a complex task, HRL decomposes the task into several easier subtasks and solve them sequentially {via MDPs} \cite{parr1998reinforcement,sutton1999between,dietterich2000hierarchical}. Recently, {HRL-based} approaches are applied to tasks like game playing \cite{kulkarni2016hierarchical,tessler2017deep}, travel planning \cite{peng2017composite}, and visual question answering and captioning \cite{wang2017video,gordon2017iqa,zhang2018multimodal}.
{Inspired by these work, given that our semantic parsing task can be naturally decomposed into 4 subtasks, we learn a two-level policy where a high-level policy decides the subtask order while a low-level policy accomplishes each subtask by asking humans clarifying questions if necessary}.

\section{Conclusion}
In this paper we explored using HRL for \textit{interactive semantic parsing}, {where an agent asks clarification questions when the initially given natural language description is ambiguous and accomplishes subtasks in an optimized order. On the If-Then recipe synthesis task, in both simulation and human evaluation\nop{simulated and real user response} settings, we have shown that our HRL agent can substantially outperform various interactive baselines \nop{and their variants }in\nop{the sense} that it produces more accurate recipes but asks the user fewer questions in general}\nop{(synthesized programs)}. {As future work, we will generalize our HRL framework to other semantic parsing tasks such as knowledge based question answering, explore better training strategies such as modeling real user noises in simulation, as well as further reduce the user interaction turns.}
\nop{As an interesting area for future work, we plan to extend the RL-based framework developed in this study to build interactive semantic parsers }
\nop{We suggest at least two interesting directions for future work: (1) The HRL framework can be generalized to resolve ambiguities in other scenarios such as knowledge-graph-based question answering \cite{yih2015semantic,berant2015agenda,chen2018sequence}, where subtasks include identifying entities, predicting relations, and associating constraints. (2) Generating context-specific agent questions to make the conversation more natural.}
\nop{The HRL framework developed in this work can be a promising direction for other scenarios such as knowledge-graph-based question answering \cite{berant2013semantic,fader2014open,yih2015semantic,berant2015agenda}.}

\section*{Acknowledgments}
We would like to thank Chris Brockett, Michel Galley and Yu Su for their insightful comments on the work. This research was sponsored in part by the Army Research Office under cooperative agreements W911NF-17-1-0412, NSF Grant IIS1815674, Fujitsu gift grant, and Ohio Supercomputer
Center \cite{OhioSupercomputerCenter1987}. The views and conclusions contained herein are those of the authors and should not be interpreted as representing the official policies, either expressed or implied, of the Army Research Office or the U.S. Government. The U.S. Government is authorized to reproduce and distribute reprints for Government purposes notwithstanding any copyright notice herein.
\nop{This research was sponsored in part by the Army Research Office under cooperative agreements W911NF-17-1-0412, NSF Grant IIS-1815674, NSF Grant CNS-1513120, Fujitsu gift grant, and Ohio Supercomputer Center \cite{OhioSupercomputerCenter1987}. The views and conclusions contained herein are those of the authors and should not be interpreted as representing the official policies, either expressed or implied, of the Army Research Office or the U.S. Government. The U.S. Government is authorized to reproduce and distribute reprints for Government purposes notwithstanding any copyright notice herein.} 

\fontsize{9.5pt}{10.5pt} \selectfont
\bibliography{interactive_sp}
\bibliographystyle{aaai}

\appendix
\section{User Answer Understanding Module} \label{app:user_ans_understand}
We define the user answer understanding module in Figure \ref{fig:policy_function} based on a bidirectional Recurrent Neural Network with attention. Specifically, for the $j$-th word in the user answer $d_i$, we concatenate its forward and backward hidden states (i.e., $h_{d_{i,j}} = [\overrightarrow{h}_{d_{i, j}}; \overleftarrow{h}_{d_{i, j}}]$) as its semantic representation. Attention weights $w_{att_j}$ over all words are computed with a trainable context vector $c$, i.e., $w_{att_j} = \text{softmax}(c^Th_{d_{i, j}})$, which will help the agent identify important words in $d_i$. User answer $d_i$ is then represented as $v_{d_i} = \sum_{j}w_{\text{att}_j}h_{d_{i,j}}$.

\section{Training Algorithm for HRL} \label{sec:train_algo}
In our work, policy functions are all updated via Stochastic Gradient Ascent.
Given the objective Eq (\ref{eq:4}), we deduce the updates for the high-level policy function below:
\begin{equation} \label{eq:6}
\nabla_\theta J(\theta) = E_{\pi^h} [\nabla_\theta \log{\pi^h(g;s)} u_{t}],
\end{equation}
where $u_{t} = \sum_{\kappa=0}^\infty \gamma^\kappa r^h(s_{t+\sum_{n=1}^\kappa N_n}, g_{t+\sum_{n=1}^\kappa N_n})$ is the summation item within the two brackets in Eq (\ref{eq:4}).

Similarly, at low level, for the ongoing subtask $g_t \in \mathcal{G} = \{st_1, st_2, st_3, st_4\}$ at time step $t$, the updates of its policy function $\pi_{g_t}^l$ are given by:
\begin{equation} \label{eq:7}
\nabla_{\phi_{g_t}} J_{g_{t}}(\phi_{g_t}) = E_{\pi_{g_t}^l} [\nabla_{\phi_{g_t}} \log{\pi_{g_t}^l (a;s)} u_{g_t,t}],
\end{equation}
where $u_{g_t,t} = \sum_{k \geq 0} \gamma^k r_{g_{t+k}}^l(s_{t+k}, a_{t+k})$.

Algorithm \ref{alg:training} details the entire training procedure. Line 1-2 initialize each policy network. Particularly, each low-level policy network is pre-trained via supervised learning (Section \ref{sec:experiments}). We train the agent on $M$ episodes (i.e., recipes). At the beginning of each episode, the state vector $s_{st_i}^l$ is initialized by calculating Eq (\ref{eq:1}-\ref{eq:2}) with $s_{st_{-i}}^l = \overrightarrow{0}$ (Line 7). Gradients of parameters $\theta$ and $\phi_{st_i}$ are accumulated during every episode. To improve efficiency, we perform gradient update for every 64 episodes (Line 31-34).

\begin{algorithm*}
  \begin{algorithmic}[1]
  \State Initialize parameters $\theta$ of the high-level policy network randomly.
  \State Initialize parameters $\phi_{st_i} \; (i=1,2,3,4)$ of each low-level policy network with supervised pre-training.
  \State Initialize gradients: $d\theta \leftarrow 0$, $d\phi_{st_i} \leftarrow 0$.
  \For{$\#episode=1$ to $M$}
    \State Reset the user simulator and get a recipe description $\mathcal{I}$.
    \State Initialize $b_i \leftarrow 0, d_i \leftarrow \emptyset, \forall i=1,2,3,4$. Observe $s_0=\{\mathcal{I}, b_i, d_i\}$.
    \State Calculate $s_{st_i}^l, \forall i=1,2,3,4$, according to Eq (\ref{eq:1}-\ref{eq:2}), with $s_{st_{-i}}^l = \overrightarrow{0}$.
    \State $global\_turn \leftarrow 1$.
    \State $t \leftarrow 0$.
    \While{$s_t$ is not terminal \textbf{and} $global\_turn \leq Max\_Global\_Turn$}
        \State Sample a subtask $g_t \sim \pi^h(g; s_t)$ according to Eq {(\ref{eq:3})}. We denote $g_t$ as $st_{i_t}$, i.e., the $i_t$-th subtask in the subtask set $\mathcal{G} = \{st_1, st_2, st_3, st_4\}$.
        \State $t_{start} \leftarrow t$.
        \State $local\_turn \leftarrow 1$, $a_t \leftarrow \varnothing$.
        \While{($a_t == \varnothing$ \textbf{or} $a_t ==$ ``AskUser'') \textbf{and} $local\_turn \leq Max\_Local\_Turn$}
             \State Sample a primitive action $a_t \sim \pi_{st_{i_t}}^l(a; s_t)$.
            \State Execute and receive a low-level reward $r_{st_{i_t}}^l(s_t, a_t)$.
            \State $d_{i_t} \leftarrow d_{i_t} \cup$ retrieved simulated user answer.
            \State Observe new state $s_{t+1}$ with new $d_{i_t}$. 
            \State Update $s_{st_{i_t}}^l$ according to Eq (\ref{eq:1}-\ref{eq:2}).
            \State $g_{t+1} \leftarrow g_t$.
            \State $t \leftarrow t + 1$.
            \State $local\_turn \leftarrow local\_turn + 1$.
        \EndWhile
        \State $d\phi_{st_{i_t}} \leftarrow d\phi_{st_{i_t}} + $ accumulated gradient according to Eq (\ref{eq:7}).
        \State Receive a high-level reward $r^h(s_{t_{start}}, g_{t_{start}})$\nop{$r_{t_{start}}^{high}$}.
        \State $b_{i_t} \leftarrow 1$.
        \State $t \leftarrow t + 1$. Observe state $s_t$.
        \State $global\_turn \leftarrow global\_turn + 1$.
    \EndWhile
    \State $d\theta \leftarrow d\theta + $ accumulated gradient according to Eq (\ref{eq:6}).
    \If{$\#episode \; \% \; 64 \; == 0$}
        \State Perform update of $\theta$ and $\phi_{st_i} \; (i=1,2,3,4)$ by gradient ascent using $d\theta$ and $d\phi_{st_i}$.
        \State $d\theta \leftarrow 0$, $d\phi_{st_i} \leftarrow 0$.
    \EndIf
  \EndFor
  \end{algorithmic}
\caption{Learning algorithm for HRL}
\label{alg:training}
\end{algorithm*}

\nop{\section{Implementation Details}
\label{app:imp}
HRL-based agents are trained for around 7 million recipes until converging. To accelerate the training of HRL-based agents, we initialize their low-level policy functions with the well-trained LAM-sup model. Particularly for HRL, we apply an ``action mask'' at high level, so that the completed subtask will not be repeatedly chosen as the next subtask. However, the agent does not rely on this mask; after being fully trained, it behaves the same as with no mask. To reduce the time on validation, we perform one validation check every 2,000 iterations, which contains 1,000 recipes randomly sampled from the complete validation set. The best parameter setting is the one that obtains the most high-level reward during the validation check.}

\section{Implementation Details}
\label{app:imple_detail}
To accelerate the training of HRL-based agents, we adopt a similar skill as \cite{gordon2017iqa,wang2017video} to initialize their low-level policy functions with the well-trained LAM-sup model. Particularly for HRL, we apply an ``action mask'' at the high level, so that the completed subtask will not be repeatedly chosen as the next subtask. However, the agent does not rely on this mask after being fully trained. The validation check is performed every 2K iterations on a randomly sampled 1K validation recipes. The best parameter setting is the one that obtains the most high-level reward in validation.

\section{User Simulator}
\label{app:user_sim}
For a trigger/action function, we design the following strategies to simulate user answers: 
\begin{enumerate}[noitemsep,leftmargin=*,topsep=0pt]
\item Revised function description: Each function has an official description on IFTTT.com, e.g., for function \texttt{Create a link note}, its description is ``\textit{This action will create a new link note from a source URL in the notebook you specify}.'' We revised the function description to make a simulated user answer, e.g., ``\textit{Create a new link note from a source URL in the notebook I specify}.''
\item Paraphrases of \nop{what does revised function name mean?} the revised function description: Specifically, we replaced words and phrases with their paraphrases, which were collected and marked as semantically ``Equivalence'' by the PPDB paraphrase database~\cite{pavlick2015ppdb}.
\item Extractions from a large collection of real-user recipe descriptions in our training set~\cite{ur2016trigger}: The extractions are based on six manually defined templates and will embrace language variation in simulated user answers. For example, for a recipe description ``If X then Y,'' X will be considered as an answer to questions about the ground-truth trigger function and Y as an answer to those about the ground-truth action function.
\end{enumerate}

\section{User Evaluation Examples}
\label{app:examples}
We present three user evaluation examples in Table \ref{tab:user_eval_examples}.
\begin{table}[ht]
\centering
\small
\begin{tabularx}{\columnwidth}{|X|l}
\hline
\multicolumn{1}{|c|}{Example 1}\\\hline
\textbf{User}: ``pinterest post to wordpress'' \\
\textbf{Ground-truth recipe}: [tc: \texttt{Feed}, tf: \texttt{New feed item}, ac: \texttt{WordPress}, af: \texttt{Create a photo post}]\\
\textbf{HRL agent}: ``Which event triggers the action?'' \\
\textbf{User}: ``when there's a new pinterest post'\\
\textbf{HRL agent}: ``Which event results from the trigger?''\\
\textbf{User}: ``create a post with photo''\\
\textbf{Agent Prediction}: [tc: \texttt{Feed}, tf: \texttt{New feed item}, ac: \texttt{WordPress}, af: \texttt{Create a photo post}] \\
\hline
\hline
\multicolumn{1}{|c|}{Example 2}\\\hline
\textbf{User}: ``foto do instagram'' \\
\textbf{Ground-truth recipe}: [tc: \texttt{Instagram}, tf: \texttt{Any new photo by you}, ac: \texttt{Email}, af: \texttt{Send me an email}]\\
\textbf{HRL agent}: ``Which event triggers the action?'' \\
\textbf{User}: ``I share new ins photo''\\
\textbf{HRL agent}: ``Which channel should act per your request?''\\
\textbf{User}: ``email''\\
\textbf{Agent Prediction}: [tc: \texttt{Instagram}, tf: \texttt{Any new photo by you}, ac: \texttt{Email}, af: \texttt{Send me an email}] \\
\hline
\hline
\multicolumn{1}{|c|}{Example 3}\\\hline
\textbf{User}: ``send fiverr orders to wunderlist'' \\
\textbf{Ground-truth recipe}: [tc: \texttt{Fiverr}, tf: \texttt{New order received}, ac: \texttt{Gmail}, af: \texttt{Send an email}]\\
\textbf{HRL agent}: ``Which event triggers the action?'' \\
\textbf{User}: ``a new order is recievied''\\
\textbf{Agent Prediction}: [tc: \texttt{Fiverr}, tf: \texttt{New email in inbox from search}, ac: \texttt{Gmail}, af: \texttt{Send an email}] \\
\hline
\end{tabularx}
\caption{Two examples (Example 1-2) from user evaluation where the HRL agent correctly interpreted the user instructions and two (Example 3) where it did not. The agent failed when there is a typo in the word ``recievied''.}
\label{tab:user_eval_examples}
\end{table}

\end{document}